\documentclass[conference]{IEEEtran}
\IEEEoverridecommandlockouts
\usepackage{cite}
\usepackage{amsmath,amssymb,amsfonts}
\usepackage{algorithmic}
\usepackage{graphicx}
\usepackage{textcomp}
\usepackage{multirow,booktabs}
\usepackage{stfloats,boldline}
\usepackage{tablefootnote}
\usepackage{tabto}
\usepackage{arydshln}
\usepackage[ruled,vlined]{algorithm2e}
\usepackage{bm}
\usepackage{caption}
\usepackage{subfig}
\usepackage{xcolor}
\def\BibTeX{{\rm B\kern-.05em{\sc i\kern-.025em b}\kern-.08em
    T\kern-.1667em\lower.7ex\hbox{E}\kern-.125emX}}

\newif\ifstatus
\statustrue

\begin{document}

\title{KORSAL: Key-point Detection based Online Real-Time Spatio-Temporal Action Localization}

\author{
\IEEEauthorblockN{Kalana Abeywardena, Shechem Sumanthiran, Sakuna Jayasundara, Sachira Karunasena,
\\
Ranga Rodrigo, and Peshala Jayasekara}

\thanks{
All authors are with the Department of Electronic and Telecommunication Engineering, University of Moratuwa, Sri Lanka (email: kalanag@uom.lk, shechems@uom.lk, sakuna@ieee.org, sachirarkarunasena@gmail.com, ranga@uom.lk, peshala@uom.lk).

This work is supported by Accelerating Higher Education Expansion and Development (AHEAD) project.}
}

\maketitle

\begin{abstract}
Real-time and online action localization in a video is a critical yet highly challenging problem.
Accurate action localization requires utilization of both temporal and spatial information.
Recent attempts achieve this by using computationally intensive 3D CNN architectures or highly redundant two-stream architectures with optical flow, making them both unsuitable for real-time, online applications.
To accomplish activity localization under highly challenging real-time constraints, we propose utilizing fast and efficient key-point based bounding box prediction to spatially localize actions. We then introduce a tube-linking algorithm that maintains the continuity of action tubes temporally in the presence of occlusions. 
Further, we eliminate the need for a two-stream architecture by combining temporal and spatial information into a cascaded input to a single network, allowing the network to learn from both types of information.
Temporal information is efficiently extracted using a structural similarity index map as opposed to computationally intensive optical flow. 
Despite the simplicity of our approach, our lightweight end-to-end architecture achieves state-of-the-art frame-mAP of \textbf{\emph{74.7\,\%}} 
on the challenging UCF101-24 dataset, demonstrating a performance gain of \textbf{\emph{6.4\,\%}}
over the previous best online methods. We also achieve state-of-the-art video-mAP results compared to both online and offline methods. Moreover, our model achieves a frame rate of \textbf{\emph{41.8}} FPS, which is a \textbf{\emph{10.7\%}} improvement over contemporary real-time methods.
\end{abstract}


\section{Introduction}
\label{sec:intro}

Spatio-temporal (ST) action localization is the task of classifying an action being performed in a series of video frames while localizing it in both space and time. Action localization is highly challenging when performed online in real-time such as in autonomous driving \cite{sahaRPN,pengRPN}. 
Actions must be localized without having access to future information to be used in online settings, while each video frame should be processed individually without using a buffer of frames in order to achieve real-time performance.

Action localization has been performed mainly through traditional methods (e.g., dense trajectories) \cite{villegas2019image, rodriguez2008action} and deep learning-based methods using Convolutional Neural Networks (CNNs) \cite{sahaRPN,pengRPN,gkioxari2015finding,weinzaepfel2015learning}. 
Due to the robust feature learning nature---as opposed to hand-crafted features in traditional methods---, CNN-based deep-learning methods for ST action localization have surpassed the traditional methods both in terms of accuracy and efficiency.   

CNN-based deep learning methods for ST action localization use two approaches: using 3D CNN frameworks that process a video as a 3D block of pixels \cite{YOWO}; or using popular object
detectors in two-stream 2D CNN frameworks with temporal linking of frame-wise detections \cite{zhang2020}. 
While superior results may be obtained by processing the entire video at once, it is impossible to use these systems for online applications.
Contemporary work that focuses on online and real-time deployment \cite{ROAD} temporally link detections using an algorithm that is unable to maintain the continuity of action tubes online. 
They also use 2-stream architectures and require expensive computation of optical flow (OF), which only marginally improves performance, but at the cost of real-time inference speed. 

In this paper, we propose an online, real-time ST action localization network by utilizing efficient key-point detection \cite{Objects_as_points} for spatial action localization, doing away with traditional anchor-box based detection architectures that require manual hyper-parameter tuning and extensive post-processing \cite{SSD,RCNN}. We do not rely on computationally intensive OF to extract explicit temporal information, rather we introduce an efficient scheme to obtain sufficient temporal information by using the structural similarity (SSIM) index map between two consecutive frames. 
We also demonstrate that the two-stream architecture is highly redundant, and superior results can be obtained by using a single network allowing it to learn only the \emph{required} temporal and spatial features for action localization.


Further, we introduce an improved tube-linking algorithm that leverages only past information. 
It extrapolates tubes for a short period in the absence of suitable detections from the object detector.
Using \cite{ROAD} as a benchmark, our proposed architecture achieves state-of-the-art frame-mean average precision (f-mAP) and video-mAP (v-mAP) for real-time, online ST action localization on UCF101-24 and J-HMDB-21 while retaining real-time speeds. 
In summary, we make the following contributions:
\vspace{0.1in}
\begin{itemize}[noitemsep]
    \item  We utilize key-point based detection architecture for the first time for the task of ST action localization, which reduces model complexity and inference time over traditional anchor-box based approaches.
    \item We demonstrate that the explicit computation of OF is unnecessary, and that the SSIM index map obtains sufficient inter-frame temporal information. 
    \item We show that the highly redundant two-stream architecture is unnecessary by providing a single network with both spatial and temporal information, and allowing it to extract necessary information through discriminative learning.
    \item We introduce an efficient tube-linking algorithm that extrapolates the tubes for a short period of time using past detections for real-time deployment.  
\end{itemize}

\section{Related Works}
\label{sec:relworks}

\subsection{Spatio-temporal action localization}

There are two approaches to ST action localization: 
    3D video processing, which processes either a sequence of frames or the entire video at once, 
    and frame-based linking techniques, which attempt to spatially localize actions within a frame, and then link those actions in the temporal domain. 

Traditional 3D video processing approaches include 3D sub-volume methods such as ST template matching \cite{rodriguez2008action}, a 3D boosting cascade \cite{ke2005efficient}, and ST deformable 3D parts \cite{tian2013spatiotemporal}. 
Recently, these have been outperformed by the 3D CNNs \cite{YOWO} that process the videos as clips in an offline fashion and localize the action in time and space.
While 3D methods are able to produce good results, they inherently suffer from being highly computationally expensive, making them unsuitable for real-time applications.

Alternatively, ST action localization can be achieved by maximizing a temporal classification path of 2D boxes detected on static frames \cite{gkioxari2015finding, weinzaepfel2015learning, tran2013video, yu2015fast}, or by searching for the optimal classification result with a branch and bound scheme \cite{yuan2011discriminative}. 
Recent works use existing 2D CNN object detection architectures to localize actions spatially and linking them temporally. To capture temporal dependencies when producing the bounding boxes, two-stream architectures have been introduced to process both the video frame and the OF concurrently, offline \cite{sahaRPN, pengRPN, gkioxari2015finding}. This method requires running two independent CNNs in parallel, leading to a two-fold increase in resource consumption. Further, it underestimates the ability of a neural network to combine spatial and temporal information to extract required semantic information through discriminative learning. Each network is restricted to spatial or temporal information only, not allowing for productive combination of both types of representations.
We demonstrate that such a separation is redundant, and better performance can be achieved by using a single network that integrates both types of information. 

Almost all mentioned 2D approaches rely on computationally expensive OF to extract temporal information between frames \cite{sahaRPN, gkioxari2015finding, Zhang2016, zhang2020}. While OF provides very accurate motion estimation \cite{horn1981determining}, there is no reason to believe that such fine-grain temporal information is necessary or even useful for a CNN to learn required temporal representations from a scene. We introduce a simple, efficient alternative to OF to obtain inter-frame temporal information -- the SSIM index map (SS-map) -- leading to increased speed, reduced complexity and improved performance, as demonstrated by our results.

Among offline action localization methods, there is limited work that focuses on real-time action detection and classification \cite{zhang2020, Zhang2016}.
However, these approaches employ large, inefficient object detection algorithms and slow OF techniques \cite{Zhang2016}. Further, they utilize future information and cannot be used for online applications \cite{zhang2020}. 
Work on online real-time ST action localization is very limited \cite{ROAD}.
A two-stream architecture that takes video frames and traditional OF as inputs to a standard CNN object detection algorithm and fuses detections prior to real-time tube generation is utilized by \cite{ROAD}. 
However, the employed linking algorithm interpolates between disjoint sets of detections that are close in time, assuming that the object detector has missed the detections in between. Therefore, it is unable to maintain tube continuity in real-time. We address this shortcoming by extrapolating forward for short periods of time, to improve tube continuity for online applications.

\subsection{Key-point detection for localization}

Although single-stage object detectors are common in ST action detection, key-point detection based action localization has not been used.  
Anchor-based single-stage object detectors \cite{SSD, YOLOV3} have a high computational complexity due to the large number of anchor box proposals. 
Further, the heuristic design of these proposals introduces many design choices, which is undesirable. 
Recently, key-point based object detection has proven to produce competitive results \cite{Objects_as_points, Cornernet, Centernet}. Detection and encoding of the top-right and bottom-left corners, followed by a matching algorithm was introduced in \cite{Cornernet}. 
Some works detect more key points per object: both corners and the center point are localized in \cite{Centernet}. 
However, the performance of such algorithms suffer from having to match sets of key-points. Single key-point detection offers the best performance with reduced complexity by detecting the centers of the object and regressing the size of the bounding box \cite{Objects_as_points}. Our real-time algorithm requires a simple and fast approach to localize actions in space.
Therefore, to the best of our knowledge, we are the first to leverage the benefits of key-point based detection for the task of spatial action localization.

\section{Proposed Architecture}
\label{sec:method}

\begin{figure*}[!t]
\centering
\includegraphics[width=\textwidth]{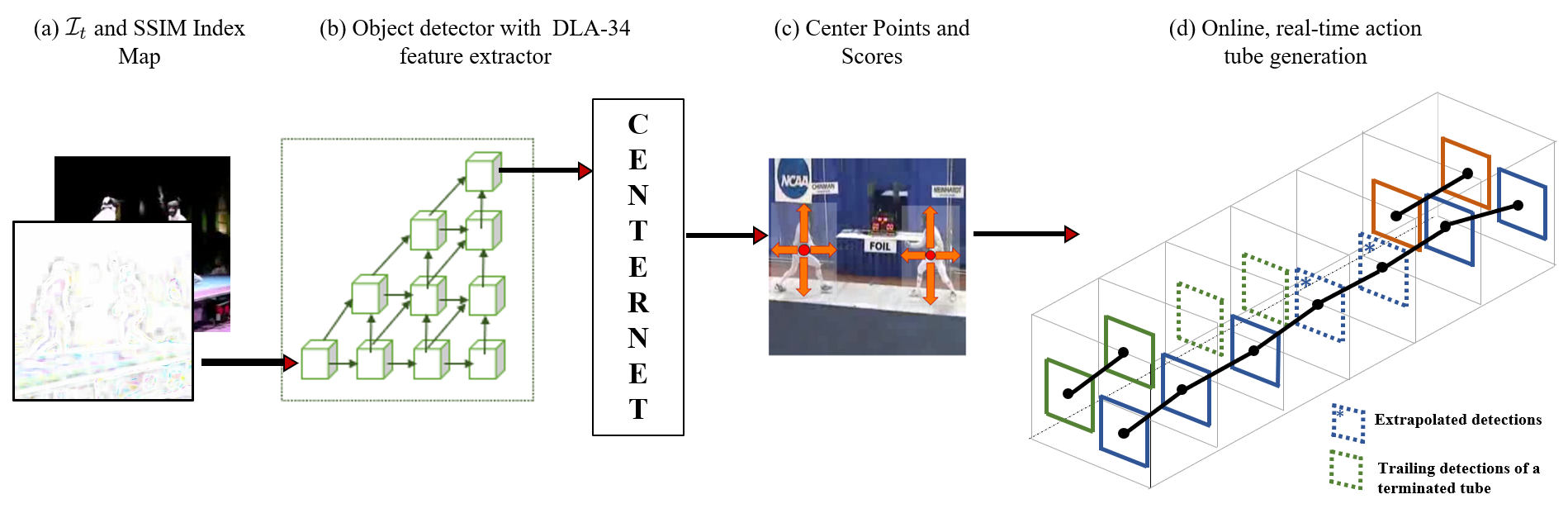}%
\caption {a) Input is $\mathcal{I}_t $ and the SS-map between $\mathcal{I}_{t-1} $ and $\mathcal{I}^*_t $ (b) DLA-34 used for feature extraction and CenterNet used to create heatmaps (\S\ \ref{ss:keydetctor}). (c) Key-points are used to build bounding boxes. (d) Acion tubes built using detections and extrapolated when required (\S\ \ref{ss:OATG}). }
\label{fig:architecture}
\end{figure*}

We propose a novel end-to-end action localization scheme by leveraging innovative temporal information incorporation, state-of-the-art key-point based object detection, and a simple, intuitive linking algorithm. A summary of our scheme is as follows.
To localize the actions occurring in a given frame of a video sequence, we utilize only that frame and the frame immediately preceding it. We shift the current frame by one pixel in all directions and use the SSIM index to determine the shifted frame which is most similar to the previous frame. 
We then concatenate the current frame and the full multi-channel SS-map between the selected shifted frame and the previous frame along the channel-axis. This is used as the input for feature extraction, for which we use the DLA-34 \cite{DLA} backbone followed by CenterNet object detection head \cite{Objects_as_points} to localize activities in the given frame using detected key-points.
The bounding boxes for each action are then passed into the linking algorithm which incrementally links them to previously existing actions to form action tubes. 
The overall architecture is presented in Fig.~\ref{fig:architecture}. 

\subsection{Temporal information representation}

\begin{figure*}[!h]
    \centering
    \includegraphics[width=\textwidth]{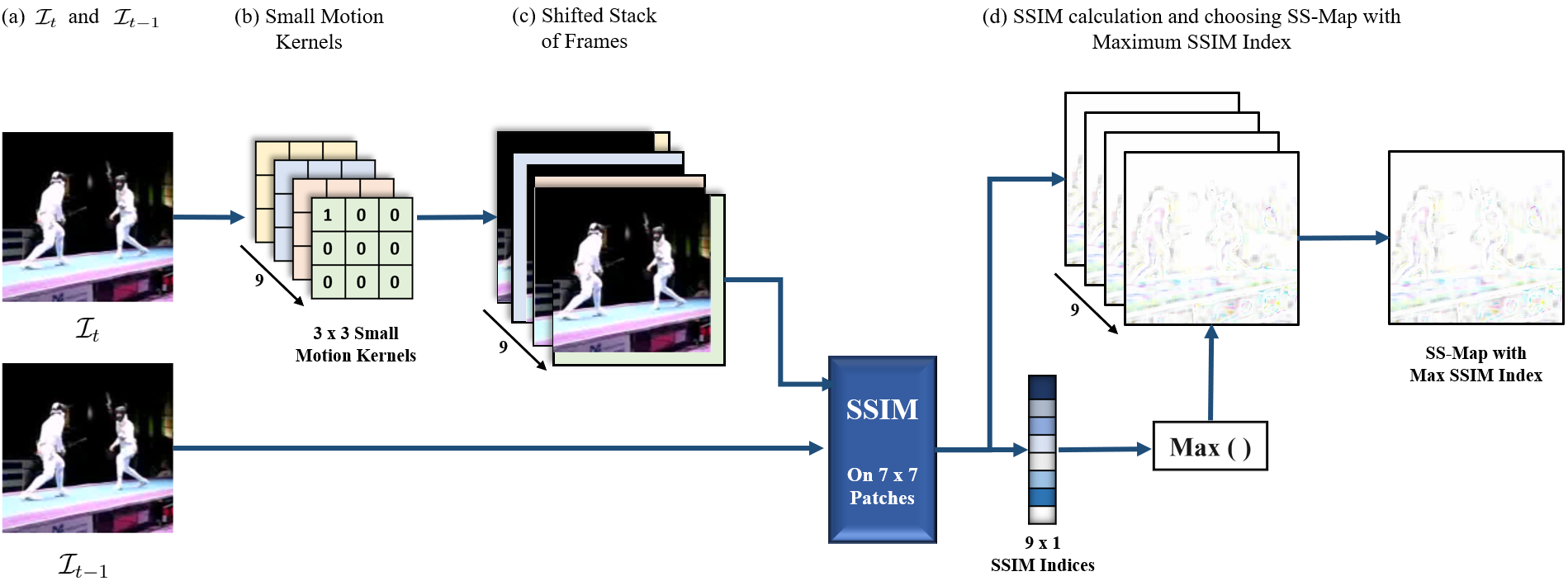}
    \caption {Obtaining temporal information using SS-map}
    \label{fig:SSIMPreprocessor}
\end{figure*}

We obtain a representation of temporal information between consecutive frames using the following 2-step procedure. Fig.~\ref{fig:SSIMPreprocessor} demonstrates the extraction method.\\

\noindent \textbf{Small motion candidate selection: }
Let the current frame be denoted by $\mathcal{I}_t$, and the previous frame be denoted by $\mathcal{I}_{t-1}$. 
To compensate for any camera motion, we shift $\mathcal{I}_t$ by one pixel in all 8 possible directions to obtain $\{\mathcal{I}_{t}^{1}, \dots, \mathcal{I}_{t}^{8}\}$. Then, from the candidates $\{\mathcal{I}_{t}, \mathcal{I}_{t}^{1}, \dots, \mathcal{I}_{t}^{8}\}$, the candidate that is most similar to $\mathcal{I}_{t-1}$ in the SSIM sense is denoted as $\mathcal{I}^*_{t}$. 

With a sufficient frame rate, the camera motion between two consecutive frames is assumed to be small enough such that a single pixel shift provides a simple and efficient way to warp the current image to the previous image.

\noindent \textbf{SS-map extraction: }
Obtaining accurate OF is computationally prohibitive for real-time applications. To capture the temporal information, we replace OF with the SS-map. The SSIM index between two images, $a$ and $b$, is defined as \cite{wang2004image}
\begin{equation}
    \mathcal{S}(a,b) = 
    \left(
    \frac{2  \mu_a \mu_b + C_1}{\mu_a^2 + \mu_b^2 + C_1}
    \right)
    \left(
    \frac{2  \sigma_{ab} + C_2}{\sigma_a^2 + \sigma_b^2 + C_2}
    \right)
    \label{SSIM}
\end{equation}
where $\mu$ and $\sigma$ refer to the sample mean and sample variance. $C_1$ and $C_2$ are small constants used to ensure stability. In order to account for local variations of structure, following \cite{wang2002universal} we apply Eq. \eqref{SSIM} over local image patches of size $7 \times 7$. This produces the SS-map. 

The SS-map between $\mathcal{I}^*_{t}$ and $\mathcal{I}_{t-1}$ is computed using Eq. \eqref{SSIM} which enables us to extract relevant temporal information of the objects of interest in the scene. The input to the feature extractor is the concatenated SS-map and $\mathcal{I}_t$, allowing the network access to both spatial and short term temporal information.

\subsection{Spatial localization of actions using key-points}
\label{ss:keydetctor}
We use a key-point based single-stage 2D CNN object detection network to localize actions within a frame. 
The network is end-to-end trainable and follows the architecture presented in \cite{Objects_as_points}. 
For a $W \times H \times C$ input, the detection network outputs a $\frac{W}{R} \times \frac{H}{R} \times N$ heatmap, where $N$ is the number of action classes, and $R$ is the down-sampling ratio. 
The heatmaps indicate the likelihood of the occurrence of center points of actions at each location. 
Additionally, the network also outputs a $\frac{W}{R} \times \frac{H}{R} \times 2$ map to indicate the width and height of actions located at each point, and a $\frac{W}{R} \times \frac{H}{R} \times 2$ map which indicates an offset from the down-sampled map to the actual center of the action bounding box. 
This leads to a greatly simplified approach over traditional anchor box-based detection architectures with reduced model complexity and inference time, which are critical for real-time applications. 

\subsection{Online action-tube generation}
\label{ss:OATG}
We present a tube-linking algorithm that matches frame-level detections obtained at time $t$, $\mathcal{D}^t = \{D_1^t, D_2^t \dots\}$, to existing action tubes generated based on detections upto time $t - 1$, $\mathcal{T}^{t-1} = \{T_1^t, T_2^t, \dots\}$. 
A detection $D_i^t$ has a bounding box $b^t_{D_i}$ and action class scores $\bm{s}^t_{D_i} \in \mathbb{R}^{C \times 1}$. $D_i^t$ can be assigned to a pre-existing tube $T_j^{t - 1}$ of class $c_{T_j}$ and score $s^{t-1}_{T_j}$ given that it has been assigned to no other tube, and $b^t_{D_i}$ has a minimum spatial overlap $\lambda$ with the most recent bounding box $b_{T_j}^{t - 1}$ in the tube. 
From the set of possible matches, similar to \cite{ROAD}, the linking algorithm greedily selects the best match for an action tube. 
Our algorithm allows unassigned detections to spawn new action tubes, and it extrapolates unassigned tubes for a maximum of $k$ frames. This ensures that tubes may begin at any point, and are not terminated due to occasional false-negative detections.

\noindent \textbf{Tube generating algorithm: }
At time $t = 1$, we select the top $n$ frame-level detections for a class $c$ by using non-maximum suppression (NMS), and each selected detection initializes an action tube of class $c$. 
At every time $t$ thereafter, having obtained the current frame-level detections $\mathcal{D}^t$, the tubes are processed in descending order of their mean score. 
For each tube $T_j^{t-1}$, we match the detection with the highest score for the tube's class $c_{T_j}$ that has a minimum intersection-over-union (IoU) ratio of $\lambda$. The tube's class is updated based on the energy maximization optimization used in \cite{ROAD}. Once this detection has been assigned, it can no longer be assigned to any other tubes. 

\noindent \textbf{Bounding box extrapolation: }
The presence of the partial occlusions and jitter in the video frames at time $t$ can cause missed detections.
Since the tube generation is performed incrementally, this may result in a discontinuity in action tubes. Therefore, if no suitable matches are found for an action tube, we maintain it for a maximum of $k$ time steps by assigning the same class confidence score as the most recent detection.

We also investigate the prediction of the extrapolated bounding box location $b^t_{T_j}$ for the tube at time $t$ without an assigned detection
using a simple motion prediction scheme.
Although this is a natural extension to extrapolation, our results do not indicate any significant performance increase, possibly due to noisy initial detections and the simplistic approach that we use. 

\begin{algorithm}[t!]
\footnotesize
\SetAlgoLined
\KwIn{
$\mathcal{T}^{t-1}$,
$\mathcal{D}^t$,
$c$,
$\lambda$,
$k$
}
\KwOut{$\mathcal{T}^{t}$} 
  \For{$T^{t-1}_j \in \mathcal{T}^{t-1}$}{
    $s \leftarrow 0; \hspace{0.1cm} m \leftarrow 0$\;
      \For{$D^t_i \in \mathcal{D}^t$}{
        \lIf {{\fontfamily{qcr}\selectfont IoU($b^t_{D_i}$, $b^{t-1}_{T_j}$)} $\geq \lambda$ {\bf and} $s < \bm{s}^t_{D_i}(c)$}{
           $b^t_{T_j} {\leftarrow}\ b^t_{D_i}; \hspace{0.1cm}\tau \leftarrow 0\; \hspace{0.1cm}s {\leftarrow}\ \bm{s}^t_{D_i}; \hspace{0.1cm}m {\leftarrow}\ i$}}
             \If{$m = 0$ {\bf and} $\tau < k$}{
                \lIf {$\fontfamily{qcr}\selectfont box\_pred = True$}{
                  $b^t_{T_j} \leftarrow$ {\fontfamily{qcr}\selectfont predict\_bbox($b^{t-1}_{T_j}$, $b^{t-2}_{T_j}$)}
                  }
                 \lElse{
                 $b^t_{T_j} \leftarrow b^{t-1}_{T_j}$
                 }
                  $\tau \leftarrow \tau + 1$\;
             }

  $s^{t}_{T_j}$, $c_{T_j} \leftarrow$ {\fontfamily{qcr}\selectfont update\_label($s^{t-1}_{T_j}$, $\bm{s}^t_{D_m}$)}\;
  }
 \caption{Online tube generation}
 \label{algo:tubealgo}
\end{algorithm}

\section{Experimental Results}
\label{sec:results}
We describe our experimental results and compare them with state-of-the-art offline and online methods that use either RGB or both RGB and OF inputs (\S\ \ref{results:comparison}). Further, for comparison we present results on action localization using only the appearance (A) information extracted by a single frame.
The results of our proposed method presented in 
Table~\ref{tab:ucfresults} and Table~\ref{tab:jhmdbresults} demonstrate that we are able to achieve state-of-the-art performance.



\noindent\textbf{Datasets:} We evaluate our framework on two datasets, UCF101-24 and J-HMDB-21. 
\textbf{UCF101-24} is a subset of UCF101 \cite{soomro2012ucf101} with ST labels, having 3207 untrimmed videos with 24 action classes, that may contain multiple instances for the same action class. \textbf{J-HMDB-21} is a subset of the HMDB-51 dataset \cite{Jhuang:ICCV:2013} having 928 temporally trimmed videos with 21 actions, each containing a single action instance.   

\noindent \textbf{Implementation: } Our keypoint-based detector is pretrained on MSCOCO object detection dataset \cite{MSCOCO}. 
We use an input image size of $256 \times 256$ for increased inference speed, and train the model for 150K iterations with a batch size of 8 on a single NVIDIA RTX 2080 Ti GPU. For the feature extractor to be robust to slight variations, instead of selecting the candidate with the highest SSIM  during training, we randomly select one of the top 3 candidates.

\noindent \textbf{Evaluation Metrics: } We evaluate our model on spatial action localization by using f-mAP with an IoU threshold of 0.5. We also report the v-mAP at several IoU thresholds to compare the performance of our model on ST action localization with the state-of-the-art results. Both metrics are computed as defined in \cite{weinzaepfel2015learning}.

\begin{figure*}[!ht]
\centering
\captionsetup[subfigure]{justification=centering}
\subfloat[Dense OF]{
        \centering
        \includegraphics[width=0.3\textwidth]{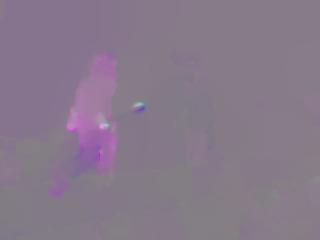}
        \label{fig:first_sub}
    }
    \hfill
\subfloat[Fast OF]{
        \includegraphics[width=0.3\textwidth]{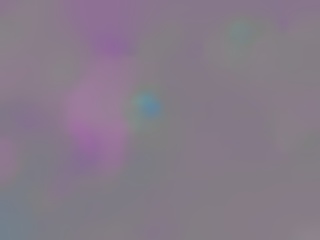}
        \label{fig:second_sub}
    }
    \hfill
\subfloat[SS-map using the selected candidate ]{
        \includegraphics[width=0.3\textwidth]{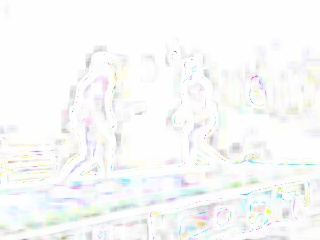}
        \label{fig:third_sub}
    }
    \vspace{0.5em}
    \captionof{figure}{Comparison between motion information extraction methods. 
    }
    \label{fig:SS-map comparison}
\end{figure*}

\subsection{Comparison with the state-of-the-art}
\label{results:comparison}
Fig. \ref{fig:SS-map comparison} compares the SS-map against the optical flow obtained by \cite{brox2004high} and \cite{kroeger2016fast} on a sample of the UCF101-24 \cite{soomro2012ucf101} dataset. The visualizations illustrates the SS-map captures not only motion information but also variations due to the other factors such as lighting and structural differences. Thus, the SS-map is rich with information about the relative differences between two consecutive frames compared to optical flow while being efficient.

Results on UCF101-24 are reported in Table~\ref{tab:ucfresults}. 
Our method surpasses the benchmark \cite{ROAD} with a marked improvement at high IoU thresholds even with the fusion of RGB and OF for the online and real-time localization of the actions while maintaining an inference speed of 41.8 FPS. Our method achieved an f-mAP score of 74.7\%, which is an improvement of 6.4\% over any other work in Table \ref{tab:ucfresults}.  

\begin{table}[!t]
    \footnotesize
    \centering
    \caption{\small{ST action localization results (v-mAP) on UCF-101-24 dataset. Last two columns compare the f-mAP and FPS. }}
    \begin{tabular}{p{2.7cm} >{\centering\arraybackslash}m{0.44cm} >{\centering\arraybackslash}m{0.45cm} >{\centering\arraybackslash}m{0.45cm} >{\centering\arraybackslash}m{0.68cm} >{\centering\arraybackslash}m{0.6cm} >{\centering\arraybackslash}m{0.39cm}}
    \hlineB{2}
        \multirow{2}{*}{\textbf{Method}} &  \multicolumn{4}{c}{\textbf{v-mAP}} & \multirow{2}{*}{\shortstack[l]{\textbf{f-mAP}\\ @0.5}} & \multirow{2}{*}{\textbf{{FPS}}}  \\
         & 0.2 & 0.5 & 0.75 & 0.5:0.95 \\
         \hline
         \noalign{\vskip 1.2pt}  
         Saha \textit{et al.}\cite{sahaRPN}$^{\diamond}$ &66.6 &36.4 &7.9 &14.4 &- & 4 \\
         Peng(w/ MR) \textit{et al.}\cite{pengRPN}$^{\diamond}$ &72.9 &- &- &- &65.7 & -\\ 
         Zhang \textit{et al.}\cite{zhang2020}$^{\diamond \star}$ &74.8 &46.6 &16.7 &21.9 &67.7 & 37.8 \\
         \noalign{\vskip 1.2pt}
         \hdashline
         \noalign{\vskip 1.2pt}
         ROAD (w/ AF) \cite{ROAD}$^{\ddagger}$ &73.5 &46.3 &15.0 &20.4 &- & 7\\
         ROAD (w/ RTF) \cite{ROAD}$^{\ddagger \star}$ &70.2 &43.0 &14.5 &19.2 &- & 28  \\
         \textbf{Our (A+AF)}$^{\ddagger}$ & 72.9& \textbf{46.7}& \textbf{16.2}& \textbf{20.9}& \textbf{70.8} & 7.7 \\
         \textbf{Our (A+RTF)}$^{\ddagger \star}$ & 69.6 & 42.1 & 15.5 & 19.3 & 69.6 & \textbf{37.9} \\
         \noalign{\vskip 1.2pt}
         \hdashline
         \noalign{\vskip 1.2pt}  
         ROAD (A) \cite{ROAD}$^{\dagger \star}$ & 69.8 &40.9 &15.5 &18.7 & - & 40\\
         \textbf{Ours (A)}$^{\dagger \star}$ & \textbf{70.2}& \textbf{44.3}& \textbf{16.6}& \textbf{20.6}& 71.8 & \textbf{52.9}\\
         \hdashline
         \textbf{Ours}$^{\dagger \star}$ & \textbf{72.7}& \textbf{43.1}& \textbf{16.8}& \textbf{20.2}& \textbf{74.7} & \textbf{41.8} \\
         \hlineB{2}
    \end{tabular}
    {$^\diamond$ Offline\ \ \  $^{\star}$ Real-time\ \ \ $^{\dagger}$ Online with no OF\ \ \ $^{\ddagger}$ Online with OF}
    \label{tab:ucfresults}
\end{table}

On J-HMDB-21, our proposed architecture outperforms all the other online, real-time models, \cite{ROAD} and offline models \cite{sahaRPN,pengRPN} by a large margin at high IoU thresholds. However, the proposed method was not able to match the results produced by our own (A) only method at the low thresholds. 
As J-HMDB-21 is a smaller dataset with only 40 frames per video, the model is unable to generalize to extract motion features from cascaded frame inputs, which might be solved by pretraining on a large action detection dataset. This contrasts with the larger UCF101-24, for which the proposed method outperformed all others. Both our methods obtained competitive accuracy with the fastest detection speed of 41.8 FPS. Additionally, our method achieves state-of-the-art results at higher IoU threshold values when compared to other offline competitors.

Unlike \cite{YOWO, zhang2020} that combine future information with causal information by processing through costly 3D CNN architectures or by using optical flow, our model is resource efficient while achieving state-of-the-art results for action detection using only causal information in a real-time manner. The inclusion of temporal information using SS-map as the candidate to the cascaded input has improved the performance for the UCF101-24 dataset both in terms of v-mAP as well as f-mAP surpassing the current state-of-the-art \cite{ROAD} with Real-Time OF. Due to the challenge of extracting combined features from the RGB frame and SS-map, our method tends to perform well with large datasets and struggles to improve beyond the appearance-only model of ours but still achieves vert competitive mAP scores.

\begin{table}[!t]  
    \footnotesize
    \centering
    \caption{\small{ST action localization results (v-mAP) on J-HMDB-21 dataset. Last two columns compare the f-mAP and FPS. }}
    \begin{tabular}{m{2.7cm} >{\centering\arraybackslash}m{0.44cm} >{\centering\arraybackslash}m{0.45cm} >{\centering\arraybackslash}m{0.45cm} >{\centering\arraybackslash}m{0.68cm} >{\centering\arraybackslash}m{0.6cm} >{\centering\arraybackslash}m{0.39cm}}
    \hlineB{2} 
      \multirow{2}{*}{\textbf{Method}} &  \multicolumn{4}{c}{\textbf{v-mAP}} & \multirow{2}{*}{\shortstack[l]{\textbf{f-mAP}\\ @0.5}}  & \multirow{2}{*}{\textbf{{FPS}}}  \\
         &0.2& 0.5& 0.75 & 0.5:0.95 \\
         \hline
         \noalign{\vskip 1.2pt}  
         Saha \textit{et al.}\cite{sahaRPN}$^{\diamond}$ &72.6 &71.5 &43.3 &40.0 &- & 4\\
         Peng(w/ MR) \textit{et al.}\cite{pengRPN}$^{\diamond}$ &74.3 &73.1 &- &- &58.5 & -\\
         Zhang \textit{et al.}\cite{zhang2020}$^{\diamond \star}$ &- &- &- &- &37.4 & 37.8 \\
         \noalign{\vskip 1.2pt}
         \hdashline
         \noalign{\vskip 1.2pt}
         ROAD (w/ AF) \cite{ROAD}$^{\ddagger}$ & 70.8 &70.1 & 43.7 &39.7 &-& 7  \\
         ROAD (w/ RTF) \cite{ROAD}$^{\ddagger \star}$ &66.0 & 63.9 &35.1 &34.4 & -& 28 \\
         
         \textbf{Our (A+AF)}$^{\ddagger}$& 68.8& 67.6& \textbf{49.9}& \textbf{43.7}& 46.9 & 7.7 \\
         \noalign{\vskip 1.2pt}
         \hdashline
         \noalign{\vskip 1.2pt}  
         ROAD (A) \cite{ROAD}$^{\dagger \star}$ & 60.8 & 59.7& 37.5& 33.9 & - & 40\\
         \textbf{Ours (A)}$^{\dagger \star}$& 59.3 & 59.2 &\textbf{48.2} &\textbf{41.2} & \textbf{51.2} & \textbf{52.9}  \\
         \hdashline
         \textbf{Ours}$^{\dagger \star}$ & \textbf{58.9} & \textbf{58.4} & \textbf{49.5} & \textbf{40.6} & \textbf{50.5} & \textbf{41.8} \\
         \hlineB{2}
    \end{tabular}
    {$^\diamond$ Offline\ \ \  $^{\star}$ Real-time\ \ \ $^{\dagger}$ Online with no OF\ \ \ $^{\ddagger}$ Online with OF}
    \label{tab:jhmdbresults}
\end{table}
\section{Conclusion and Future Works}
\label{sec:conclusion}

In this paper, we proposed a method to solve the challenging problem of \emph{online} and \emph{real-time} spatio-temporal action localization by utilizing simple and efficient key-point based detection architectures.
We further improved upon existing linking algorithms to maintain temporal continuity by extrapolating the future positions of action tubes to compensate for missed detections online.
We showed that the pre-computation of OF to capture motion information affects real-time performance, and we integrated temporal and appearance feature extraction into a single network.
We demonstrated that our approach is able to run faster and achieve better performance than state-of-the-art methods on the UCF101-24 and J-HMDB-21 datasets. 

Further extensions of this work can explore an integrated tube-linking algorithm and faster feature extraction backbones. Moreover, temporally aware feature extraction can also be investigated.





\section{Ablation Studies}
\label{sec:ablation}

We carry out various experiments to determine the impact of each of the introduced components. We analyze the impact that the different sections of the algorithm have on the overall inference time. We investigate the effects of changing the temporal information representation method, introducing extrapolation and bounding box prediction to the linking algorithm, and using an increased frame gap between cascaded inputs.

\begin{table}[!ht]
    \footnotesize
    \centering
    \captionsetup{justification = centering}
    \caption{Inference timing analysis}
    \vspace{1em}
    \begin{tabular}{p{2.7cm} >{\centering\arraybackslash}m{0.44cm} >{\centering\arraybackslash}m{0.45cm} >{\centering\arraybackslash}m{0.68cm} >{\centering\arraybackslash}m{0.40cm} >{\centering\arraybackslash}m{0.55cm} >{\centering\arraybackslash}m{0.49cm}}
    \hlineB{2}
        \noalign{\vskip 1.5pt}
        {\textbf{Framework Module}} &  {\textbf{Ours}} & {A + DSIM} & {\textbf{A + $\mathcal{I}_{t-1}$}} & {\textbf{A}} & {\textbf{RTF}} & {\textbf{A + AF}}  \\
        \noalign{\vskip 1.5pt} 
         \hline
         \noalign{\vskip 1.5pt} 
         Temporal INFO EXT (ms) & 5.0 & 5.0 & - & - & 7.0 & 110.0 \\
         Detection network (ms)
         & 16.4 & 16.4 & 16.4 & 16.4 & 16.4 & 16.4\\
        Tube generation time (ms) & 2.5 & 2.5 & 2.5 & 2.5 & 3.0 & 3.0\\
        \hlineB{2}
        \noalign{\vskip 1.5pt}
         \textbf{Overall (ms)} & 23.9 & 23.9 & 18.9 & 18.9 & 26.4 & 129.4\\
         \hlineB{2}
    \end{tabular}
    \label{tab: inference time}
\end{table}

\noindent{\textbf{Inference time: }} We analyze the inference times for different variations of our pipeline based on the different modules in the framework and the overall inference time in Table~\ref{tab: inference time}. Evidently, any preprocessing will have an impact on the inference time. Thus, the SS-map achieves a balance between the run-time and the accuracy over the other variations in the framework. 

\vspace{2mm}

\begin{table}[!t]
    \footnotesize
    \centering
    \caption{\small{Variations of temporal information representation (UCF-101-24) }}
    \begin{tabular}{p{2.7cm} >{\centering\arraybackslash}m{0.64cm} >{\centering\arraybackslash}m{0.65cm} >{\centering\arraybackslash}m{0.65cm} >{\centering\arraybackslash}m{0.78cm} >{\centering\arraybackslash}m{0.75cm}}
    \hlineB{2}
        \multirow{2}{*}{\textbf{Method}} &  \multicolumn{4}{c}{\textbf{v-mAP}} & \multirow{2}{*}{\shortstack[l]{\textbf{f-mAP}\\ @0.5}}   \\
         & 0.2 & 0.5 & 0.75 & 0.5:0.95 \\
         \hline
         \noalign{\vskip 1.2pt}  
         $\mathcal{I}_{t-1}$ &71.6 &44.1 &17.0 &20.7 &74.4 \\
         SS-map &72.4 &43.0 &16.6 &20.2 &74.7 \\ 
         DSIM index map &73.4 &44.9 &16.4 &20.7 &74.5  \\
         \noalign{\vskip 1.2pt}
         \hlineB{2}
    \end{tabular}
    
    \label{tab: temporal representation ucf}
\end{table}

\begin{table}[!t]
    \footnotesize
    \centering
    \caption{\small{Variations of temporal information representation (J-HMDB-21) }}
    \begin{tabular}{p{2.7cm} >{\centering\arraybackslash}m{0.64cm} >{\centering\arraybackslash}m{0.65cm} >{\centering\arraybackslash}m{0.65cm} >{\centering\arraybackslash}m{0.78cm} >{\centering\arraybackslash}m{0.75cm}}
    \hlineB{2}
        \multirow{2}{*}{\textbf{Method}} &  \multicolumn{4}{c}{\textbf{v-mAP}} & \multirow{2}{*}{\shortstack[l]{\textbf{f-mAP}\\ @0.5}}   \\
         & 0.2 & 0.5 & 0.75 & 0.5:0.95 \\
         \hline
         \noalign{\vskip 1.2pt}  
         $\mathcal{I}_{t-1}$ &57.2 &55.9 &48.1 &39.9 &47.9 \\
         SS-map &58.9 &58.4 &49.4 &40.5 &50.5 \\ 
         DSIM index map &56.4 &55.9 &49.2 &39.9 &49.9  \\
         \noalign{\vskip 1.2pt}
         \hlineB{2}
    \end{tabular}
    
    \label{tab: temporal representation hmdb}
\end{table}

\noindent \textbf{Temporal information representation methods: }
We investigate different representations of temporal information for our proposed model in Table~\ref{tab: temporal representation ucf} and Table~\ref{tab: temporal representation hmdb}. Apart from SS-map, we evaluate the structural dissimilarity (DSIM) index map \cite{loza2006structural} using $\mathcal{I}^*_{t}$ and using $\mathcal{I}_{t-1}$ without any preprocessing as the input along with $\mathcal{I}_{t}$. Overall, the SS-map outperforms other methods on J-HMDB21. Although the DSIM method yields the best v-mAP on UCF101-24, SS-map provides the best f-mAP results. We propose that using $\mathcal{I}_{t-1}$ achieves lower results as the SS-map provides convenient cues to the network as to which areas it should pay attention to, which is not provided when the raw previous frame is used as the second input. 
\vspace{2mm}


\begin{table}[!ht]  
    \footnotesize
    \centering
    \captionsetup{justification = centering}
    \caption{Linking algorithm variations (UCF-101-24)}
    \vspace{1em}
    \begin{tabular}{m{1.6cm} 
    >{\centering\arraybackslash}m{0.8cm} >{\centering\arraybackslash}m{0.8cm} >{\centering\arraybackslash}m{0.55cm} >{\centering\arraybackslash}m{0.58cm} >{\centering\arraybackslash}m{0.58cm} >{\centering\arraybackslash}m{0.80cm} 
    }
    \hlineB{2} 
    \noalign{\vskip 1.2pt} 
    \multirow{3}{*}{\textbf{\shortstack[l]{\textbf{Linking}\\ Algorithm}}} &  
    \multicolumn{2}{c}{\textbf{Improvement}} &
    \multicolumn{4}{c}{\textbf{UCF-101-24}} \\
    \noalign{\vskip 1.2pt} 
    \cline{2-7}
    
    & \multirow{2}{*}{EXPLT} & \multirow{2}{*}{BOXP}
    & \multicolumn{4}{c}{\textbf{v-mAP}} 
    
    \\
    & & & 0.2& 0.5& 0.75 & 0.5:0.95 \\
      
         \hline
         \noalign{\vskip 1.2pt}  
          Original &  &  & 72.6 & 43.4 & 16.8 & 20.3 \\
          Ours  & \checkmark &  & 72.7 & 43.1 & 16.8 & 20.2 \\
          Ours & \checkmark & \checkmark & 72.4 & 43.0 & 16.6 & 20.2 \\
         \noalign{\vskip 1.2pt}
         \hlineB{2}
    \end{tabular}
    \\
    \label{tab: linking variations ucf}
\end{table}

\begin{table}[!ht]  
    \footnotesize
    \centering
    \captionsetup{justification = centering}
    \caption{Linking algorithm variations (J-HMDB-21)}
    \vspace{1em}
    \begin{tabular}{m{1.6cm} 
    >{\centering\arraybackslash}m{0.8cm} >{\centering\arraybackslash}m{0.8cm} >{\centering\arraybackslash}m{0.55cm} >{\centering\arraybackslash}m{0.58cm} >{\centering\arraybackslash}m{0.58cm} >{\centering\arraybackslash}m{0.80cm} 
    }
    \hlineB{2} 
    \noalign{\vskip 1.2pt} 
    \multirow{3}{*}{\textbf{\shortstack[l]{\textbf{Linking}\\ Algorithm}}} &  
    \multicolumn{2}{c}{\textbf{Improvement}} &
    \multicolumn{4}{c}{\textbf{J-HMDB-21}} \\
    \noalign{\vskip 1.2pt} 
    \cline{2-7}
    
    & \multirow{2}{*}{EXPLT} & \multirow{2}{*}{BOXP}
    & \multicolumn{4}{c}{\textbf{v-mAP}} 
    
    \\
    & & & 0.2& 0.5& 0.75 & 0.5:0.95 \\
      
         \hline
         \noalign{\vskip 1.2pt}  
          Original &  &  & 58.8 & 58.3 & 49.4 & 40.5 \\
          Ours  & \checkmark &  & 58.9 & 58.4 & 49.4 & 40.6 \\
          Ours & \checkmark & \checkmark & 58.9 & 58.4 & 49.4 & 40.5 \\
         \noalign{\vskip 1.2pt}
         \hlineB{2}
    \end{tabular}
    \\
    \label{tab: linking variations hmdb}
\end{table}

\noindent \textbf{Analysis on Linking Algorithm Variations: }
We analyzed the proposed improvements to the linking algorithm in terms of how they affect the overall v-mAP for the two datasets in Table~\ref{tab: linking variations ucf} and Table~\ref{tab: linking variations hmdb}. EXPLT denotes extrapolation, and BOXP denotes bounding box location prediction. The results indicate that extrapolating detections for a short time improves results by compensating for missed detections. The intuitive idea of bounding box prediction during the extrapolation does not improve the results of the experiments. We therefore maintain detection locations when a tube is extrapolated. This simple scheme proves sufficient to improve performance.
\vspace{2mm}

\begin{figure}[!ht]
\centering
\captionsetup[subfigure]{justification=centering}
\subfloat[UCF101-24]{
        \centering
        \includegraphics[width=0.45\textwidth]{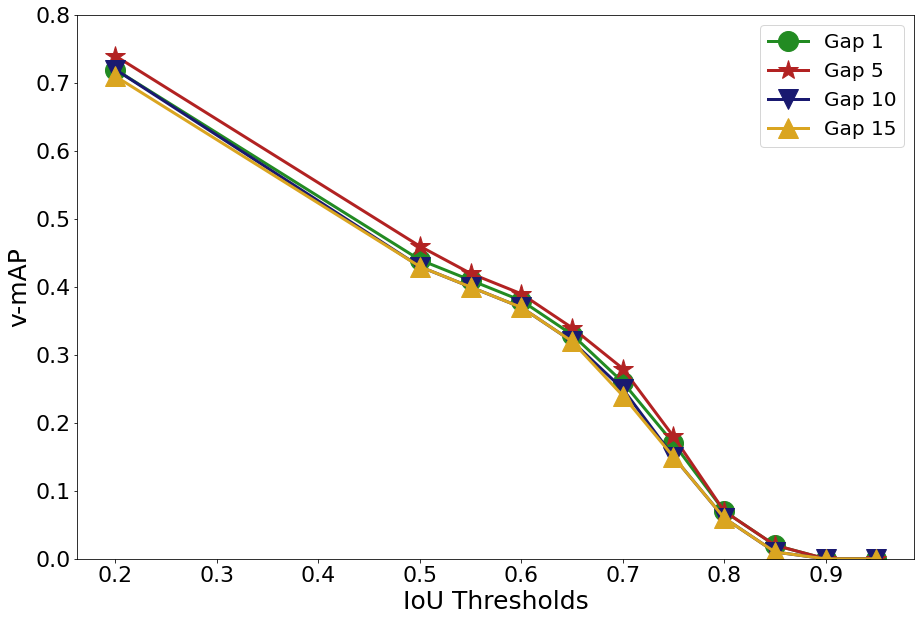}
        \label{fig:abs4_ucf}
    }
    \hfill

\subfloat[J-HMDB21]{
        \includegraphics[width=0.45\textwidth]{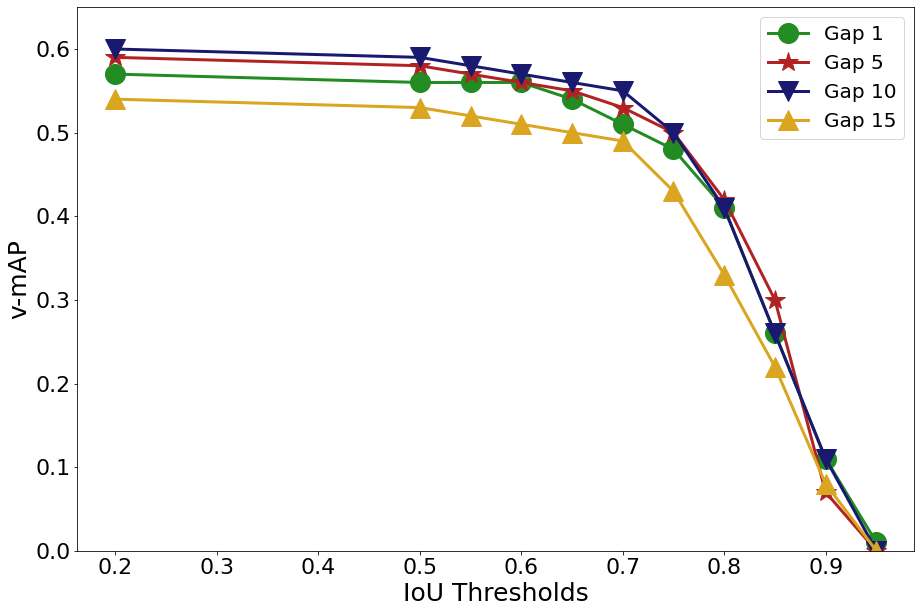}
        \label{fig:abs4_hmdb}
    }
    \hfill
    \vspace{0.5em}
    \captionof{figure}{Analysis of frame gap between the current frame and the past frame utilized.}
    \label{fig: frame gap}
\end{figure}

\noindent \textbf{Effect of Frame Gap on Motion: }
Due to high video frame rates, the difference between two consecutive frames may be negligible, thus containing little temporal information. We analyzed how action localization is impacted when varying the frame gap between the current image and the past image we use to compute the SS-map. For this we used the test setting where the input is $\mathcal{I}_t$ and $\mathcal{I}_{t-k}$, where $k$ is the frame gap we utilize. Based on Fig.~\ref{fig: frame gap}, obtaining temporal information using consecutive frames is difficult. There is a stronger information between frames which are further separated in time: for UCF24 the best results are obtained at frame gap of 5 and for J-HMDB21 at  10. This indicates that the optimal frame-gap is \emph{data dependent}. However, for both the cases the frame gap of 5 between the current and the past frame provides better results than using consecutive frames. We leave the exploration of the frame gap optimization to future work.

\bibliographystyle{IEEEtran}
\bibliography{IEEEabrv,BibTex}

\end{document}